\pdfoutput=1

\documentclass[11pt]{article}

\usepackage[preprint]{acl}

\usepackage{times}
\usepackage{latexsym}
\usepackage{algorithm}
\usepackage{algpseudocode}
\usepackage{amsmath}
\usepackage{multirow}
\usepackage{xspace}
\usepackage{amssymb}
\usepackage{booktabs}
\usepackage{xcolor}

\usepackage[T1]{fontenc}

\usepackage[utf8]{inputenc}

\usepackage{microtype}

\usepackage{inconsolata}

\usepackage{graphicx}

\usepackage{booktabs}

\definecolor{codegray}{gray}{0.9}
\newcommand{\code}[2]{%
  \begingroup\setlength{\fboxsep}{1pt}%
  \colorbox{#1}{#2}%
  \endgroup
}
\newcommand{\isscore}{\code{orange!3}{Intersectional Sensitivity}\xspace}
\newcommand{\iss}{\code{orange!3}{$IS$}\xspace}
\newcommand{\modelname}{\code{cyan!5}{BiasConnect}\xspace}
\newcommand{\mitname}{\code{purple!5}{InterMit}\xspace}
\newcommand{\hardprompt}{{\textsl{PM}}}

\title{Mitigate One, Skew Another? \\ Tackling Intersectional Biases in Text-to-Image Models}

\author{Pushkar Shukla$^{*1}$
~~~~~
Aditya Chinchure$^{*2}$
~~~~~
Emily Diana$^{3}$
~~~~~
Alexander Tolbert$^{4}$ \\
~~~~~
{\bf Kartik Hosanagar}$^{5}$
~~~~~
{\bf Vineeth N. Balasubramanian}$^{6}$
~~~~~
{\bf Leonid Sigal}$^{2}$
~~~~~
{\bf Matthew A. Turk}$^1$ \\
$^1$Toyota Technological Institute at Chicago ~~~~~ $^2$University of British Columbia \\ $^3$Carnegie Mellon University, Tepper School of Business ~~~~~ $^4$Emory University \\
$^5$University of Pennsylvania, The Wharton School ~~~~~ $^6$Indian Institute of Technology Hyderabad\\
{\tt\small \{pushkarshukla, mturk\}@ttic.edu}  ~~~~~
{\tt\small \{aditya10, lsigal\}@cs.ubc.ca}
}

\begin{document}
\maketitle
\begin{abstract}
The biases exhibited by text-to-image (TTI) models are often treated as independent, though in reality, they may be deeply interrelated. Addressing bias along one dimension—such as ethnicity or age—can inadvertently affect another, like gender, either mitigating or exacerbating existing disparities. Understanding these interdependencies is crucial for designing fairer generative models, yet measuring such effects quantitatively remains a challenge. To address this, we introduce \modelname, a novel tool for analyzing and quantifying bias interactions in TTI models. \modelname uses counterfactual interventions along different bias axes to reveal the underlying structure of these interactions and estimates the effect of mitigating one bias axis on another.  These estimates show strong correlation (+0.65) with observed post-mitigation outcomes.

Building on \modelname, we propose \mitname, an intersectional bias mitigation algorithm guided by user-defined target distributions and priority weights. \mitname achieves lower bias (0.33 vs. 0.52) with fewer mitigation steps (2.38 vs. 3.15 average steps), and yields superior image quality compared to traditional techniques. Although our implementation is training-free, \mitname is modular and can be integrated with many existing debiasing approaches for TTI models, making it a flexible and extensible solution.
\end{abstract}    
\begin{figure}[t]
  \centering
   \includegraphics[width=1.0\linewidth]{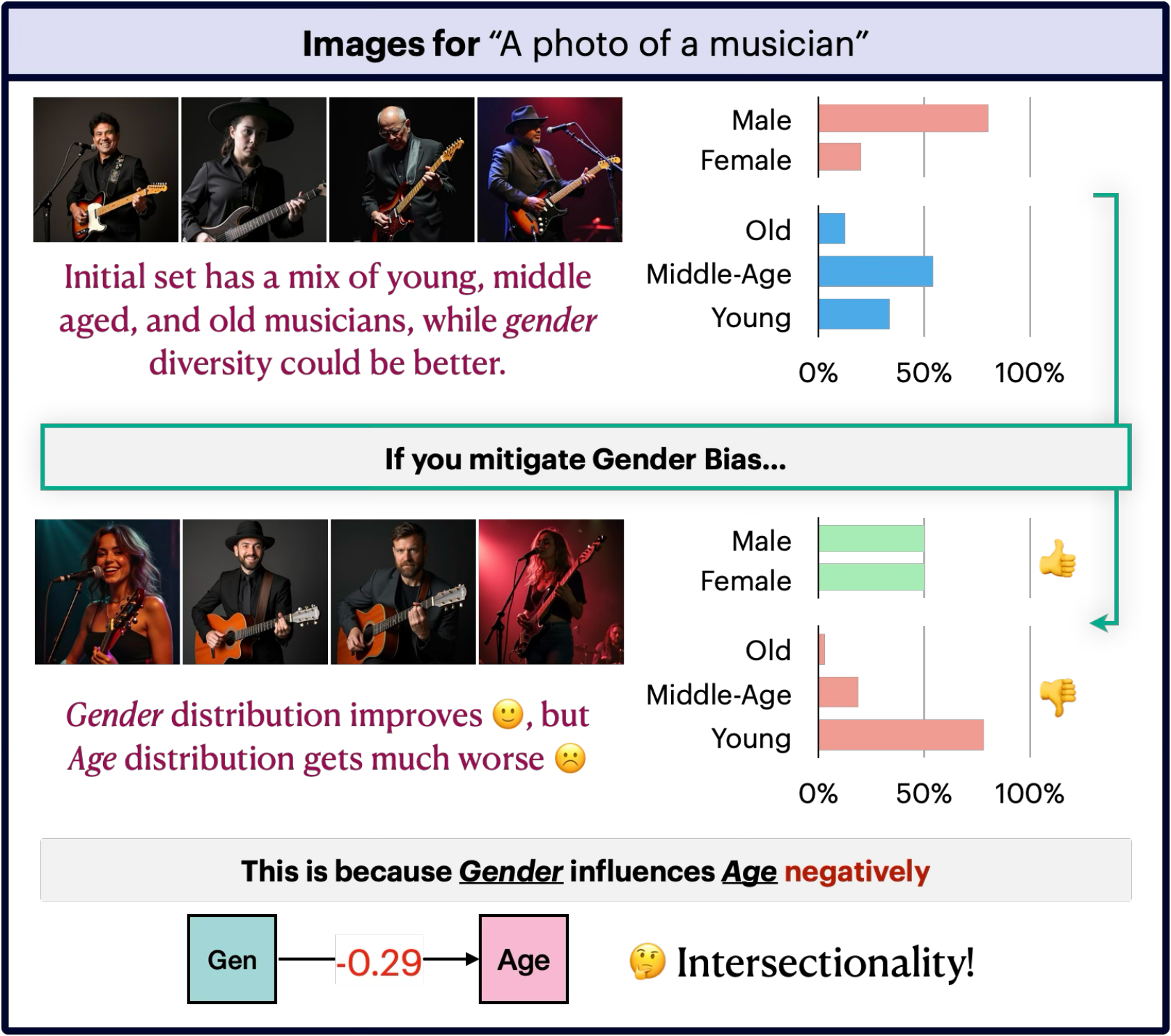}
   \caption{An example for which \modelname estimates a negative impact of bias mitigation along one axis on another axis. For this query, increasing the gender diversity (Gen) skews age distribution (Age) for images of musicians generated by Flux-dev.}
   \label{fig:header}
\end{figure}
\section{Introduction}
\label{sec:intro}

Text-to-Image (TTI) models such as DALL-E \citep{ramesh2021zero}, Imagen \citep{saharia2022photorealistic}, and Stable Diffusion \cite{rombach2022high} have become widely used for generating visual content from textual prompts. Despite their impressive capabilities, these models often inherit and amplify biases present in their training data \cite{wang_towards_2022,chinchure2024tibet,cho2023dall}. These biases manifest across multiple social  and non-social dimensions -- including 
gender, race, clothing, and age -- leading to skewed or inaccurate representations. As a result, TTI models may reinforce harmful stereotypes and societal norms \citep{bender2021dangers, birhane2021multimodal}. While significant efforts have been made to evaluate and mitigate societal biases in TTI models \citep{wang2023t2iat, cho2023dall,ghosh2023person,esposito2023mitigating,bianchi2023easily,chinchure2024tibet}, these approaches often assume that biases along different dimensions (e.g., gender and race) are independent of each other. Consequently, they do not account for relationships between these dimensions. For instance, as illustrated in Figure \ref{fig:header}, mitigating gender (male, female) may effectively diversify the gender distribution in a set of generated images, but this mitigation step may negatively impact the diversity of another bias dimension, such as age. This relationship between two bias dimensions highlights the intersectional nature of these biases.

The concept of \textit{intersectionality}, first introduced by 
Crenshaw \citep{crenshaw1989demarginalizing}, motivates the need to understand how overlapping social identities such as race, gender, and class contribute to systemic inequalities. In TTI models, these intersections can have a significant impact. As a motivating study, we independently mitigated eight bias dimensions over 26 occupational prompts on Stable Diffusion 1.4, using a popular bias mitigation strategy, ITI-GEN \cite{zhang2023iti} (see \ref{sup:intro_study}). We found that while the targeted biases were reduced in most cases, biases along other axes were negatively affected in over $29\%$ of the cases. This suggests that for an effective bias mitigation strategy, it is crucial to understand which biases are intersectional. Additionally, it is important to strive towards building a more holistic bias mitigation algorithm that can either mitigate multiple biases simultaneously or predict what biases cannot be mitigated together. 


To understand how biases in TTI models influence one another, we propose \modelname, the first analysis tool that evaluates biases while explicitly modeling their intersectional relationships. Unlike prior methods that treat biases in isolation, \modelname\ identifies how mitigating one bias can positively or negatively affect others. 
Specifically, \modelname\ uses a novel metric, the \isscore\ (\iss), to quantify how mitigation along one axis affects others. These \iss\ scores show a strong correlation (+0.65) with observed intersectional outcomes post-mitigation. We validate our approach through robustness studies and qualitative analyses, demonstrating its utility for auditing open-source TTI models. 

Furthermore, we extend \modelname\ with a holistic intersectional bias mitigation algorithm, \mitname. While we propose an effective and straightforward implementation in this paper, \mitname\ is modular and can be integrated with any existing sequential bias mitigation method. Unlike prior approaches that assume fixed ideal distributions and treat all biases equally, \mitname\ allows users to define arbitrary target distributions, select specific bias axes, and assign custom priority weights to each bias—enabling flexible joint mitigation and informed reasoning about conflicting biases.

In our evaluation, \mitname\ outperforms existing methods by mitigating biases more effectively, producing higher-quality images, and requiring fewer mitigation steps. Moreover, unlike other methods, it can handle a larger number ($>3$) of bias axes and alerts users when mitigation along one axis adversely affects others.


\begin{figure*}
  \centering
   \includegraphics[width=\linewidth]{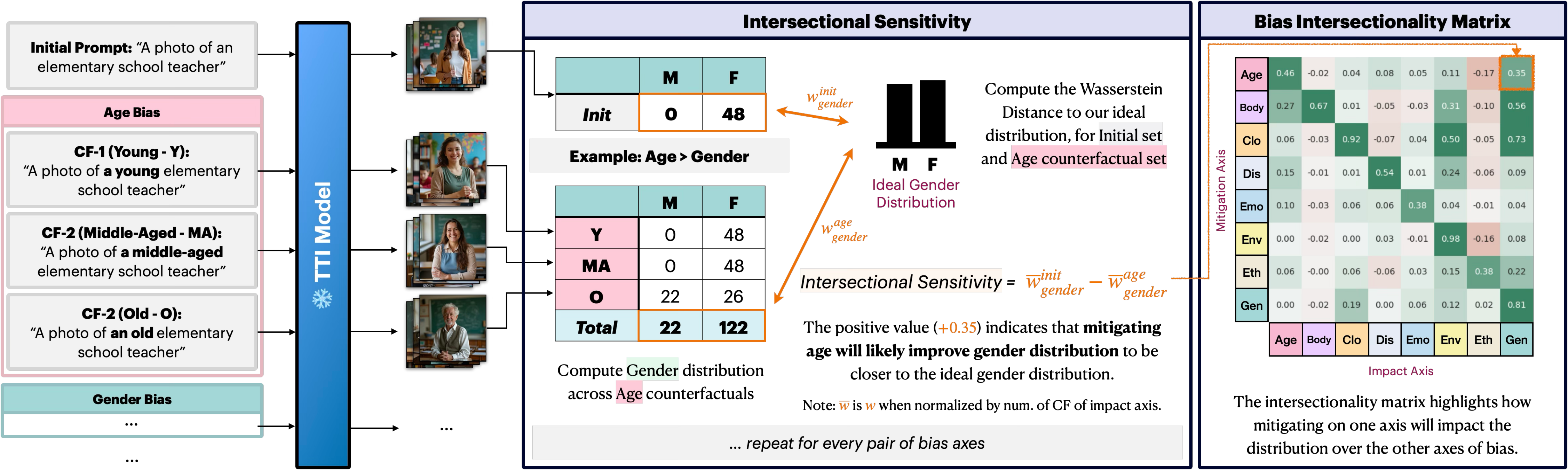}
   \caption{\textbf{An overview of \modelname}. We use a counterfactual-based approach to measure how interventions along a single bias axes impact other bias axes. Our metric \isscore   estimates how bias mitigation on one axis impacts another. Our results are visualized as a  a matrix called the Bias Intersectionality Matrix.}
   \label{fig:approach}
\end{figure*}

\section{Related Work }
\label{sec:related_work}
\subsection{Intersectionality and Bias in AI}

Intersectionality, introduced by Crenshaw \cite{crenshaw1989demarginalizing}, describes how multiple forms of oppression—such as racism, sexism, and classism—intersect to shape unique experiences of discrimination. Two key models define this concept: the additive model, where oppression accumulates across marginalized identities, and the interactive model, where these identities interact synergistically, creating effects beyond simple accumulation \cite{curry2018killing}. In the context of AI, most existing work \cite{diana2023correcting,kavouras2023fairness,kearns2018preventing} aligns more closely with the additive model, focusing on quantifying and mitigating biases in intersectional subgroups. This perspective has influenced fairness metrics \cite{diana2021minimax,foulds2020intersectional,ghosh2021characterizing} designed to assess subgroup-level performance, extending across various domains, including natural language processing (NLP) \cite{lalor2022benchmarking,lassen2023detecting,guo2021detecting,tan2019assessing} and recent large language models \cite{kirk_bias_2021,ma2023intersectional,devinney2024we,bai2025explicitly}, multimodal research \cite{howard2024socialcounterfactuals,hoepfinger2023racial}, and computer vision \cite{wang2020towards, steed2021image}. These approaches typically measure disparities across predefined demographic intersections and propose mitigation strategies accordingly. Our work aligns with the interactive model of intersectionality, using counterfactual analysis with TTI models, where we intervene on a single bias axis to assess its ripple effects on others.




\subsection{Bias in Text-to-Image Models}

Extensive research has been conducted on evaluating and mitigating social biases in both image-only models \cite{buolamwini2018gender, seyyed2021underdiagnosis, hendricks2018women, meister2023gender, wang2022revise, liu2019fair, joshi2022fair, wang2023overwriting} and text-only models \cite{bolukbasi2016man, hutchinson2020social, shah2020predictive, garrido2021survey, ahn2021mitigating}. More recently, efforts have expanded to multimodal models and datasets, addressing biases in various language-vision tasks. These investigations have explored biases in embeddings \cite{hamidieh2023identifying}, text-to-image generation \cite{cho2023dall, bianchi2023easily, seshadri2023bias, ghosh2023person, zhang2023iti, wang2023t2iat, esposito2023mitigating}, image retrieval \cite{wang2022assessing}, image captioning \cite{hendricks2018women, zhao2021scaling}, and visual question-answering models \cite{park2020fair, aggarwal2023fairness, hirota2022gender}.

Despite these advances, research on intersectional biases in TTI models remains limited. Existing evaluation frameworks such as T2IAT \cite{wang2023t2iat}, DALL-Eval \cite{cho2023dall}, and other studies \cite{ghosh2023person, bianchi2023easily, friedrich2023fair} primarily assess biases along predefined axes, such as gender \cite{wang2023t2iat, cho2023dall, esposito2023mitigating, bianchi2023easily}, skin tone \cite{wang2023t2iat, cho2023dall, ghosh2023person, esposito2023mitigating, bianchi2023easily}, culture \cite{esposito2023mitigating, wang2023t2iat}, and geographical location \cite{esposito2023mitigating}. 
While these works offer key insights into single-axis bias detection and mitigation, they lack a systematic examination of how biases on one axis influence another—a core aspect of intersectionality. The closest research, TIBET \cite{chinchure2024tibet}, visualizes such interactions, but our approach goes further by systematically quantifying bias interactions, and using these interactions for mitigation.

\section{Approach}
\label{sec:Approach}

The objective of \modelname\ is to identify and quantify the intersectional effects of intervening on one bias axis (\(B_x\)) to mitigate that bias, on any other bias axis (\(B_y\)). \modelname works by systematically altering input prompts and analyzing the resulting distributions of generated images (see Fig. \ref{fig:approach}). To achieve this, we leverage counterfactual prompts by modifying specific attributes (e.g., male and female) along a bias axis (e.g., gender) and examine how these interventions impact other bias dimensions (e.g., age and ethnicity). If modifying one bias axis through counterfactual intervention causes significant shifts in the distribution of attributes along another bias axis, it indicates an intersectional dependency between these axes. We first construct prompt counterfactuals and generate images using a TTI model (Sec. \ref{sec:Approach:Imggen}). Subsequently, to identify bias-related attributes in the generated images, we use a Visual Question Answering (VQA) model (Sec. \ref{sec:Approach:VQA}). Finally, to quantify the intersectional effects, and to identify whether these effects are positive or negative, we compute the  \isscore\ (Sec. \ref{sec:Approach:CausalEffect}). 

\subsection{Counterfactual Prompts \& Image Generation}
\label{sec:Approach:Imggen}

Given an input prompt \( P \) and bias axes \( B = \{B_1, B_2, \dots, B_n\} \), we generate counterfactual prompts 
\(\{CF_i^1, \dots, CF_i^j\} \) for each bias $B_i \in B$. These counterfactual prompts may be templated (Appendix Table \ref{tab:fulldataset}) or LLM-generated.
The original prompt \( P \) and its counterfactuals are then used to generate images with the TTI model to measure intersectional effects.

\subsection {VQA-based Attribute Extraction}
\label{sec:Approach:VQA}
 
To facilitate the process of extracting bias-related attributes from the generated images, we use VQA. This is inspired by previous approaches on bias evaluation, like TIBET \cite{chinchure2024tibet} and OpenBias \cite{d2024openbias}, where a VQA-based method was used to extract concepts from generated images. Following TIBET, we use MiniGPT-v2 \cite{chen2023minigptv2} in a question-answer format to extract attributes from generated images.

For the societal biases we analyze, we have a list of predefined questions (Appendix \ref{sup:VQA}) corresponding to each bias axis in $B$, and each question has a choice of attributes to choose from. For example, for the gender bias axis, we ask the question ``\texttt{\small [vqa] What is the gender (male, female) of the person?}''. Note that every question is multiple choice (in this example, \texttt{\small male} and \texttt{\small female} are the two attributes for gender). For datasets where counterfactuals are dynamically generated (e.g. TIBET dataset), an LLM-generated set of questions is used instead. The questions asked for all images of prompt \( P \) and its counterfactuals \( CF_i^j \) remain the same. With the completion of this process, we have attributes for all images, where each image has one attribute for each bias axis in $B$.

\subsection{Computing Intersectional Sensitivity}
\label{sec:Approach:CausalEffect}
Our objective is to understand how the impact of interventions on \( B_x \) affects \( B_y \) in a positive or negative direction concerning an ideal distribution. 
To address this, we propose a metric that quantifies the impact of bias mitigation on dependent biases with respect to an ideal distribution. 



\noindent{\bf Defining an Ideal Distribution.} We first define a desired (ideal) distribution \(D^*\), which represents the unbiased state we want bias axes to achieve. This can be a real-world distribution of a particular bias axis, a uniform distribution (which we use in our experiments), or anything that suits the demographic of a given sub-population. 

\noindent{\bf Measuring Initial Bias Deviation.} Given the images of initial prompt \(P\), we compute the empirical distribution of attributes associated with bias axis \(B_y\), denoted as \( D_{B_y}^{\text{init}} \). We then compute the Wasserstein distance between this empirical distribution and the ideal distribution:
\begin{equation}
w_{B_y}^{\text{init}} = W_1(D_{B_y}^{\text{init}}, D^*)
\end{equation}
\noindent where \( W_1(\cdot, \cdot) \) represents the Wasserstein-1 distance. The Wasserstein-1 distance (also known as the Earth Mover's Distance) between two probability distributions \( D_1 \) and \( D_2 \) is defined as:
\begin{equation}
W_1(D_1, D_2) = \inf_{\gamma \in \Pi(D_1, D_2)} \mathbb{E}_{(x,y) \sim \gamma} [|x - y|]
\end{equation}
\noindent where \( \Pi(D_1, D_2) \) is the set of all joint distributions \( \gamma(x, y) \) whose marginals are \( D_1 \) and \( D_2 \), and \( |x - y| \) represents the transportation cost between points in the two distributions.

We use $\overline{w}^{init}_{B_y}$ to measure the amount of bias in the image set, where $\overline{w}_{B_y}$ is computed by normalizing $w_{B_y}$ based on the number of counterfactuals in $B_y$. $\overline{w}^{init}_B \in [0,1]$ where 1 indicates that the distribution is completely biased and 0 indicates no bias.

\noindent{\textbf{Intervening on $B_x$.}}
Next, say we intervene on \(B_x\) to simulate the mitigation of bias $B_x$. This intervention ensures that all counterfactuals of \(B_x\) are equally represented in the generated images. For example, if \(B_x\) is gender bias, we enforce equal proportions of male and female individuals in the dataset. This intervention is in line with most bias mitigation methods proposed for TTI models, like ITI-GEN \cite{zhang2023iti}. Using our counterfactuals along \(B_x\), we sum the distributions on \(B_y\) across all counterfactuals of \(B_x\). This sum across the counterfactuals of \( B_x \) yields a new empirical distribution of \( B_y \), denoted \( D_{B_y}^{B_x} \), simulating the effect of mitigating \( B_x \) (See Fig \ref{fig:approach}). We compute its Wasserstein distance from the ideal distribution.
\begin{equation}
w_{B_y}^{B_x} = W_1(D_{B_y}^{B_x}, D^*)
\label{eqn:biasscore}
\end{equation}
\noindent{\textbf{Computing \isscore.}} To quantify the effect of mitigating \(B_x\) on \(B_y\), we define the metric, \isscore, as:
\begin{equation}
IS_{xy} = \overline{w}_{B_y}^{\text{init}} - \overline{w}_{B_y}^{B_x} 
\label{eqn:metric}
\end{equation}
\noindent as Wasserstein distance is sensitive to the number of counterfactuals, and $IS_{xy} \in [-1,1]$. A positive value (\( IS_{xy} > 0 \)) indicates that mitigating \( B_x \) improves \( B_y \), bringing it closer to the ideal distribution, while a negative value (\( IS_{xy} < 0 \)) suggests it worsens \( B_y \), moving it further from the ideal. If \( IS_{xy} = 0 \), mitigating \( B_x \) has no effect on \( B_y \). This approach enables us to assess whether addressing one bias (e.g., gender) improves or worsens another (e.g., ethnicity) in generative models, providing a systematic way to evaluate trade-offs and unintended consequences in bias mitigation strategies. 



\begin{figure*}[ht]
  \centering
   \includegraphics[width=\linewidth]{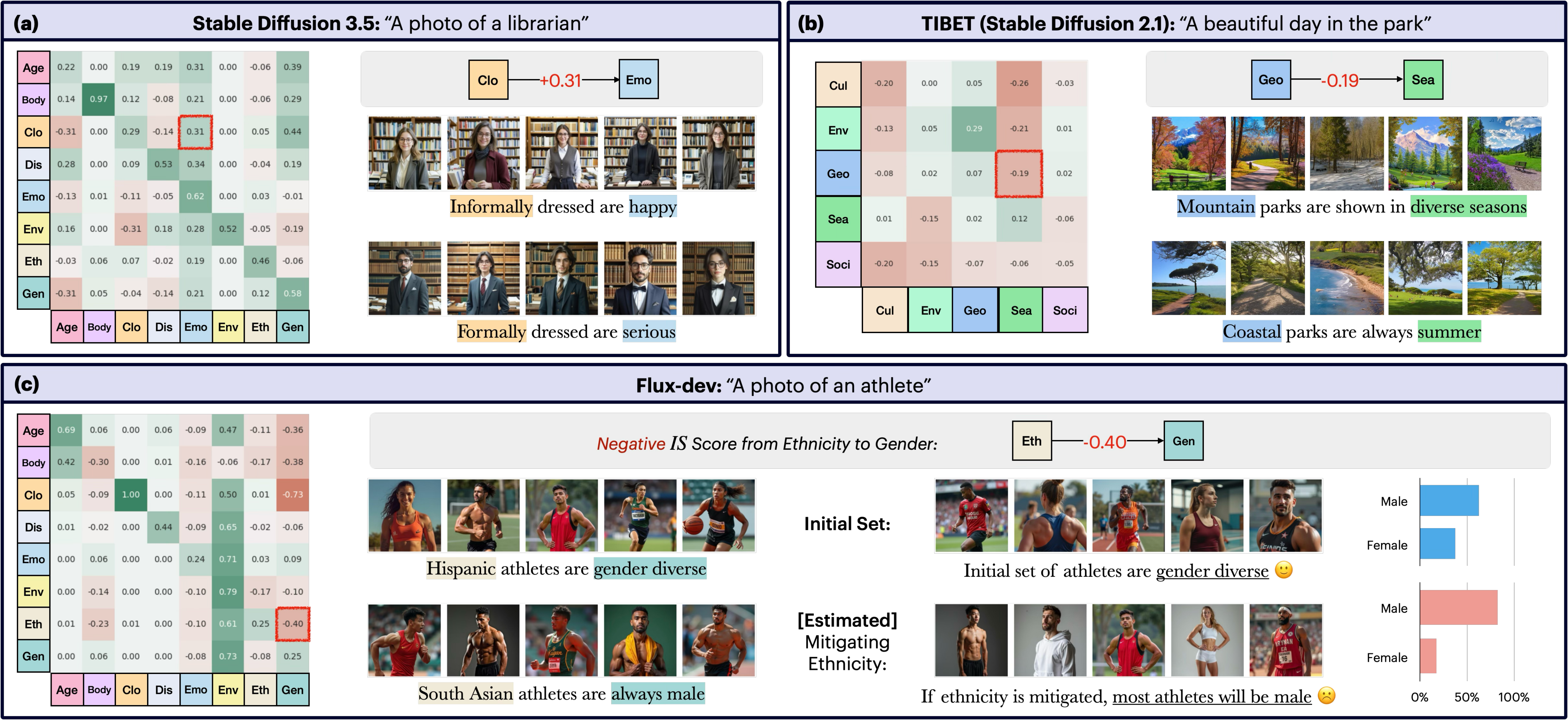}
   \caption{Analyzing bias intersectionality matrices from \modelname. (a) Shows how mitigating clothing bias also mitigates emotion bias. (b) Explores interactions between non-traditional bias axes in the TIBET dataset. (c) Reveals that generating ethnically diverse athletes reduces gender diversity. \modelname can allow the user the user to understand whether interventions along one dimension impact other dimensions positively or negatively.   }
   \label{fig:examples}
\end{figure*}

\subsection{Visualization}

To visualize $IS$ scores comprehensively, we use a \textit{Bias Intersectionality Matrix} $\mathbf{S}$, where each entry $IS_{ij}$ quantifies the effect of intervening on row $B_i$ on column $B_j$ for mitigation. This matrix captures directional dependencies and enables a structured analysis of intersectional bias effects.

\section{Intersectional Bias Mitigation using \modelname}

This section introduces an iterative strategy for mitigating intersectional biases, designed to be modular and compatible with existing sequential bias mitigation algorithms.
We propose a subjective, user-guided mitigation framework built on top of \modelname, called \mitname. The framework allows users to select a subset of bias dimensions from a predefined set and assign mitigation priorities to each. Additionally, users can specify a desired target distribution that the model should conform to, providing both flexibility and control over the mitigation process. Our proposed framework (in Algorithm~\ref{alg:intersectional_mitigation}) leverages a matrix $\mathbf{S}$ at each step to iteratively reduce bias across multiple axes. 

Given a TTI model $M$, a subset of selected axes by the user $B^* \subseteq B$,  let $\mathbf{p} \in \mathbb{R}^{|B^*|}$ be a user-defined \textit{priority vector} that encodes the relative importance of mitigating bias along each axis, where $|\mathbf{p}|_1 = 1$. The ideal desired distribution $D^*$, for each bias axes in $B^*$ is also specified by the user. Given the aforementioned information, we first calculate a bias score for initial model $M^{(0)}$ by taking the dot product of the priority vector $\mathbf{p}$ and the initial measures of biases $B^*$ ($\overline{w}^{init}_{B*}$) computed using Eq. \ref{eqn:biasscore}. This is denoted by $\tau =\langle \overline{w}^{init}_{B*},p \rangle$ and measures the overall bias of the model on $B^*$ at any timestep. We proceed to mitigation if $\tau$ is greater than a threshold $\epsilon$. 

To choose which bias axis to mitigate on, we extract the submatrix $\mathbf{S}' \in \mathbb{R}^{n \times |B^*|}$ consisting of the relevant columns from $\mathbf{S}$ obtained using \modelname. For each row $\mathbf{s}'_i$ of $\mathbf{S}'$, we compute a similarity score $\gamma_i = \langle \mathbf{s}'_i, \mathbf{p} \rangle$, which quantifies the alignment between the $i$-th intersectional bias and the desired direction of mitigation. The bias axis $i^* = \arg\max_i \gamma_i$ with the highest alignment score is selected for targeted mitigation in the current iteration. The model is then updated to reduce bias along the direction corresponding to $i^*$, using a mitigation method, giving $M^{(1)}$. After mitigation, we generate a new set of images, recompute $\tau$, and continue the mitigation process if $\tau > \epsilon$. 

\begin{algorithm}
\caption{\mitname: Intersectional Mitigation}
\label{alg:intersectional_mitigation}
\begin{algorithmic}[1]
\Require Relevant bias axes $B^* \subseteq B$, priority vector $\mathbf{p} \in \mathbb{R}^{|B^*|}$ with $\|\mathbf{p}\|_1 = 1$, sensitivity matrix $\mathbf{S} \in \mathbb{R}^{n \times {|B^*|}}$, bias threshold $\epsilon$, TTI model $M$
\Ensure Final mitigated model $M^{(t)}$ with $\tau < \epsilon$
\State Initialize model $M^{(0)}$, set iteration counter $t \gets 0$
\Repeat
    \State Extract submatrix $\mathbf{S}' \in \mathbb{R}^{n \times |B^*|}$ from $\mathbf{S}$
    \State Extract priority vector $\mathbf{p} \in \mathbb{R}^{|B^*|}$
    \For{$i = 1$ to $n$}
        \State Compute similarity score $\gamma_i \gets \langle \mathbf{s}'_i, \mathbf{p} \rangle$
    \EndFor
    \State Identify target axis: $i^* \gets \arg\max_i \gamma_i$
    \State Mitigate axis $i^*$ to update model: $M^{(t+1)}$
    \State Compute  bias score $\tau^{(t+1)} =\langle \overline{w}^{init}_{B*},p \rangle $
    \State $t \gets t + 1$
\Until{$\boldsymbol{\tau}^{(t)} < \epsilon$}
\State \Return $M^{(t)}$
\end{algorithmic}
\end{algorithm}

\section{Experiments}
\label{sec:Experiments}

We evaluate \modelname\ for its ability to study intersectional biases across multiple models and prompts (Section \ref{sec:mainintersectionality}, \ref{sec:mitigation}) and its robustness (\ref{sec:robustness}). Following that, we use \mitname\ for mitigation, and compare it to an existing strategy (\ref{sec:compareitigen}). 


\subsection{Experiment Setup}
\label{sec:ModelsDatasets}


We conduct experiments on two prompt datasets, across six TTI models:

\noindent\underline{Occupation Prompts:}  To facilitate a structured evaluation, we develop a dataset with 26 occupational prompts, along eight distinct bias dimensions: gender, age, ethnicity, environment, disability, emotion, body type, and clothing. We generate 48 images for all initial counterfactual prompts using five TTI models: Stable Diffusion 1.4, Stable Diffusion 3.5, Flux \cite{flux2024}, Playground v2.5 \cite{li2024playground} and Kandinsky 2.2 \cite{kandinsky22,razzhigaev2023kandinsky}. Further details about the prompts, bias axes, and counterfactuals are provided in the Appendix  \ref{sup:dataset}.

\noindent\underline{TIBET dataset:} 
The TIBET dataset includes 100 creative prompts with unique LLM-generated bias axes and counterfactuals \cite{chinchure2024tibet} for each prompt, helping us test with a diverse array of biases. Additionally, it provides 48 Stable Diffusion 2.1-generated images per initial and counterfactual prompt (see Appendix \ref{sup:TIBET}).

\noindent\textbf{Mitigation.} \mitname\ can use any sequential mitigation method, but we consider a simple training-free mitigation method using only prompt modifications (\hardprompt). At each mitigation step, we modify the initial prompt to introduce counterfactual concepts associated with the mitigated bias axis. Over multiple steps, we create collections of counterfactual prompts that include all permutations of all mitigated axes (see \ref{sup:hardprompt}). We emperically set $\epsilon = 0.35$ for all our experiments. To compare our method to a traditional mitigation approach, we select ITI-GEN \cite{zhang2023iti}, as it uses a similar FairToken-based permutation approach. 

\subsection{Studying prompt-level intersectionality}
\label{sec:mainintersectionality}

\begin{table*}[t]
\small
  \centering
  \begin{tabular}{l|ccccc|cc}
  \toprule 
\textbf{Method} & {\color[HTML]{F56B00} \textbf{quality}} $\uparrow$ & {\color[HTML]{F56B00} \textbf{real}} $\uparrow$ & {\color[HTML]{F56B00} \textbf{natural}} $\uparrow$ & {\color[HTML]{F56B00} \textbf{colorfulness}} $\uparrow$ & {\color[HTML]{F56B00} \textbf{IsP?}} $\uparrow$ & {\color[HTML]{00009B} \textbf{MitAmt}} $\downarrow$ & {\color[HTML]{00009B} \textbf{MitSteps}} $\downarrow$ \\
\midrule
ITI-GEN (SD1.4)  & 0.73    & 0.92   & 0.37    & 0.45     & 92.8\%    &    0.52   & 100\%  \\
\mitname-\hardprompt\ (SD1.4) & \textbf{0.82}   & \underline{0.98}  & \underline{0.58} & \underline{0.66} & \underline{99\%}  &  \textbf{0.33} & \textbf{75.6\%} (2.38/3.15)   \\
\midrule
\mitname-\hardprompt\ (SD3.5)  & 0.78    & \textbf{0.99}     & \textbf{0.92}  & \textbf{0.74}   & \textbf{100\%}    & 0.27$^*$   & 76\%$^*$ (2.71/3.57)  \\                  
\bottomrule
\end{tabular}
  \caption{\textbf{Comparing our Mitigation Algorithm to ITI-GEN}. We mitigate a randomly chosen subset of 2-5 biases for prompts in the occupation set, and compute {\color[HTML]{F56B00} visual quality} metrics and {\color[HTML]{00009B} mitigation outcomes}. We find that our algorithm uses 22\% fewer mitigation steps, while still yielding higher mitigation amount and quality. $^*$Indicates we use a different prompt set and priority on SD3.5, so these should not be compared to SD1.4 results.
 }
  \label{tab:mitigationres}
\end{table*}

\modelname\ supports prompt-level analysis of intersectional biases (Fig. \ref{fig:examples}), helping users identify key bias axes and effective mitigation strategies. For example, in Fig. \ref{fig:examples}(a), Stable Diffusion 3.5 shows a causal link between clothing and emotion—informal attire leads to happier depictions of librarians (\iss\ = 0.31), suggesting clothing changes can diversify emotional portrayal. In contrast, Fig. \ref{fig:examples}(c) shows ethnicity negatively affecting gender diversity, with South Asian athletes mostly depicted as male (\iss\ = -0.40), indicating that addressing ethnicity alone may worsen gender bias. These insights support model comparison and targeted bias mitigation through \mitname.


\subsection{Validating \isscore}
\label{sec:mitigation}
Our approach estimates how counterfactual-based mitigation affects bias scores using the \isscore. To validate this, we use ITI-GEN and \hardprompt\ to mitigate biases along each dimension, and measure the correlation between pre- and post-mitigation $IS$ values. We achieve an average correlation of \fbox{+0.65} across occupations using ITI-GEN. Certain axes like musician (+0.91), accountant (+0.81) and lawyer (+0.82) have especially high correlations. An average correlation of \fbox{+0.95} using \hardprompt\ is unsurprising, as it uses similar counterfactual prompts for mitigation. The strong correlation observed between pre- and post-mitigation bias scores suggests that our approach effectively estimates the potential impacts of bias mitigation, motivating the need to account for intersectionality in mitigation. More details on our experimental setup are in Appendix \ref{sup:mitigation}.

\begin{figure}
  \centering
   \includegraphics[width=\linewidth]{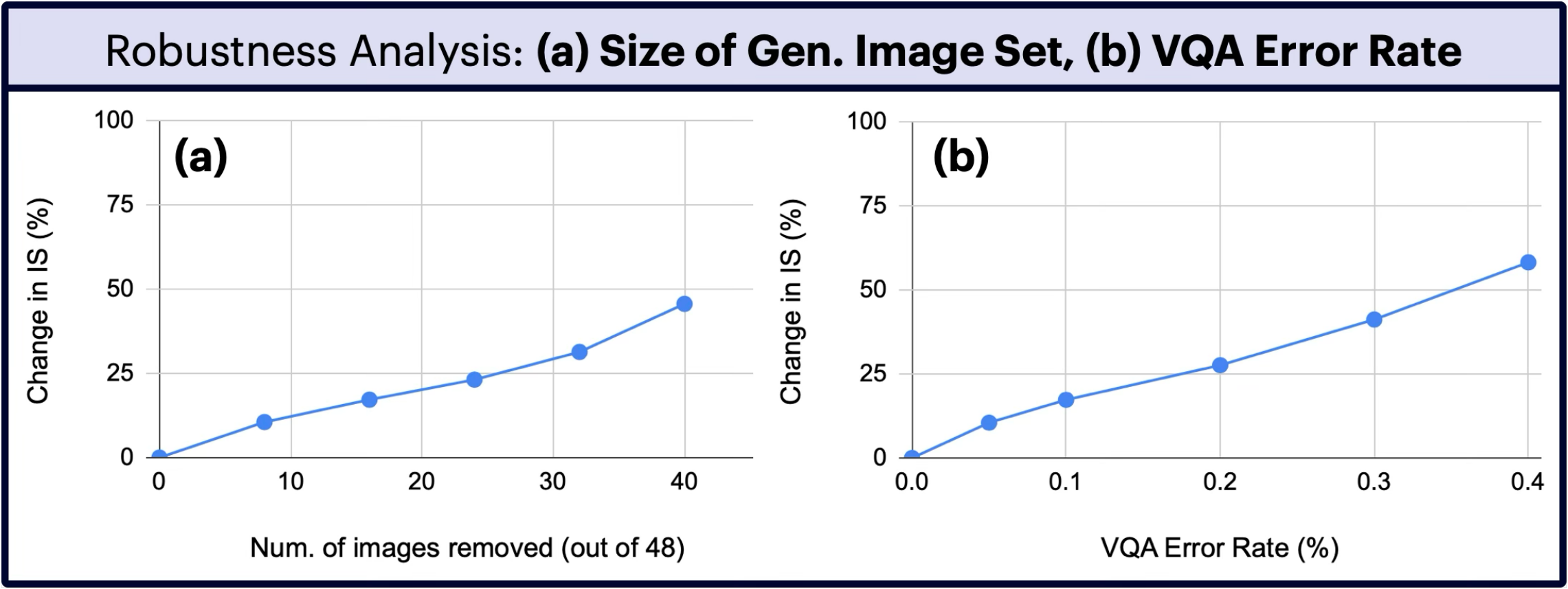}
   \caption{\textbf{Sensitivity analysis on \modelname}. We evaluate the robustness of our approach by analyzing the impact of VQA errors and the effect of the number of images on \isscore.}
   \label{fig:sensitivity}
\end{figure}

\subsection{Robustness of \modelname}
\label{sec:robustness}

We analyze the robustness of our method by evaluating the impact of  number of images (Fig. \ref{fig:sensitivity}(a))  and VQA error rate (Fig. \ref{fig:sensitivity}(b)) on \isscore\ values. Our method uses 48 images per prompt to study bias distributions. Removing 8 images (16.6$\%$) results in 10.5$\%$ change, an removing 32 images (66.6$\%$) yields a 31.3$\%$. This sub-linear impact suggests that TTI models often generate similar bias distributions (e.g., always depicting nurses as females), preserving overall trends despite fewer images. Therefore, our approach is robust to moderate reductions in image count, but very small sets of images will significantly affect \iss\ values. To test the robustness over VQA errors, we randomly change the VQA answers to a different answer (simulating an incorrect answer), from 5\% to 40\% of the time. We observe that even with low error rates of 5\% and 10\%, \iss\ values change by 10\% and 17.3\% respectively. Here, the impact is compounded twice, because an error can skew the distribution away from one counterfactual towards another, and that a 5\% error causes 13,478 answers out of a total of 269,568 answers to be changed, which is substantial. Nonetheless, we note that this impact remains linear. As VQA models improve, achieving low error rates for robustness becomes practical.

\begin{figure*}
  \centering
  \includegraphics[width=\linewidth]{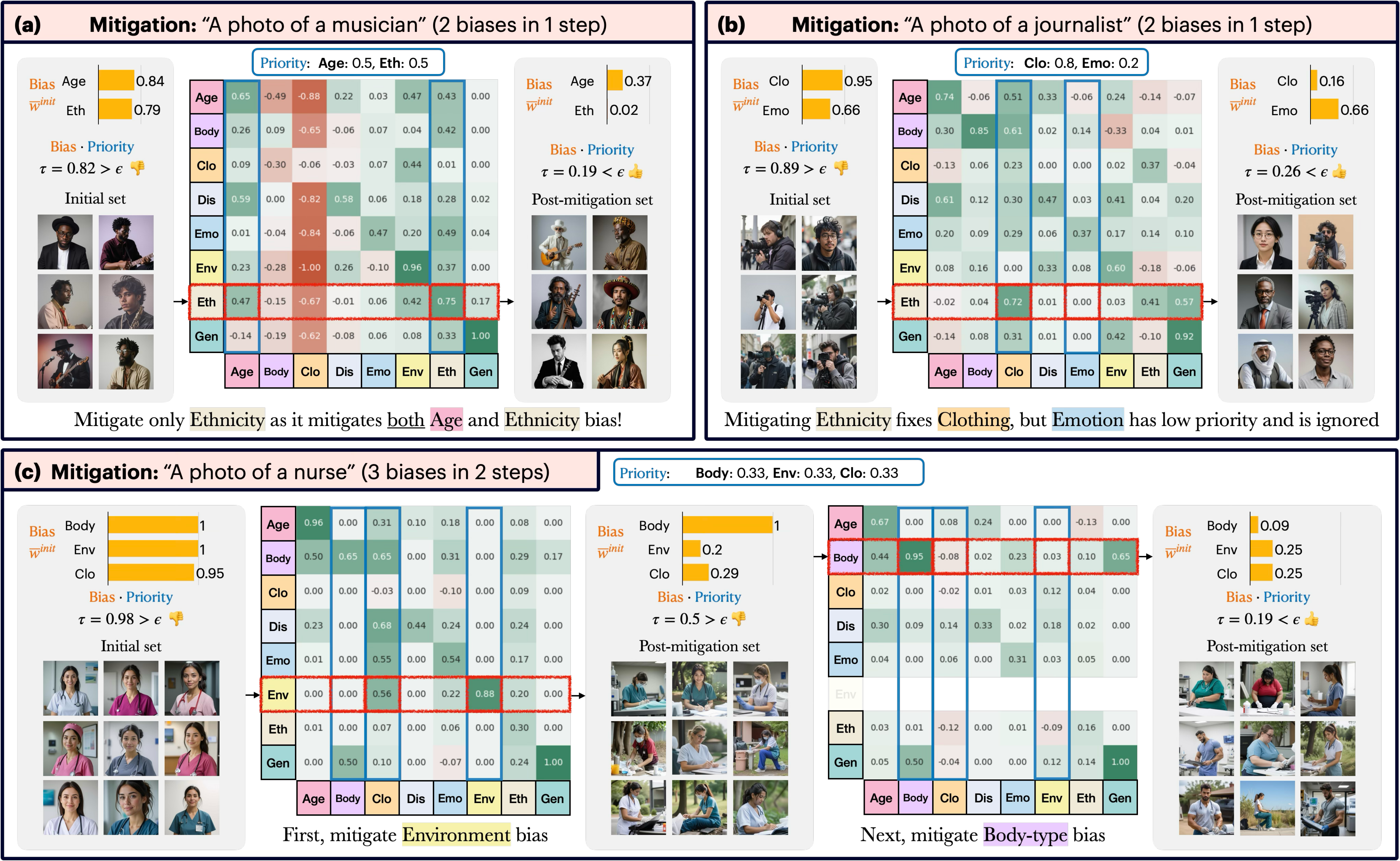}
   \caption{Three examples of mitigation using \mitname. Priority vectors guide the mitigation process. Columns that are a part of sub-matrix $\mathbf{S}'$ are in blue. As shown in (a) and (c), the algorithm mitigates multiple biases in fewer steps. (b) shows how user-defined priorities guide the process and when thresholds are met. Mitigating one axis, like ethnicity (Eth), can also affect others like clothing (Clo) and emotion (Emo), revealing bias interdependencies.   }
   \label{fig: applications}
\end{figure*}




\subsection{Analyzing \mitname\ for Mitigation}
\label{sec:compareitigen}

To evaluate the effectiveness of our bias mitigation approach, we compare it against ITI-GEN \cite{zhang2023iti}.  ITI-GEN is designed for SD1.4 and is limited in its ability to mitigate more than three axes of bias for any given prompt. We override it in our experiments to facilitate a broader comparison. In contrast, our method combines \hardprompt\ with the intersectional mitigation algorithm \mitname.

\noindent\textbf{Prompts and Metrics.} For SD1.4, we randomly select subsets of biases to mitigate, assigning equal priority to each bias, in the occupation set. For SD3.5, we use 15 occupation prompts, targeting intersectional biases with some priorities weighted. Details are in Appendix \ref{sup:mitprompts}. Table \ref{tab:mitigationres} quantifies the mitigation amount (MitAmt: averaging $\tau^T$ post-mitigation across all prompts), and efficiency (MitSteps: ratio of number of biases mitigated to the number of biases in $\mathbf{p}$). Visual quality is evaluated using CLIP-IQA metrics \cite{wang2022clipiqa} and using the VQA query: “\texttt{\small [vqa] Is there a person in the image?}” (IsP?).

\noindent\textbf{Results.} On SD1.4, \mitname-\hardprompt\ achieves significantly lower bias (0.33 vs. 0.52 for ITI-GEN), while requiring only 75.6\% of the steps (ITI-GEN will always mitigate all biases in the priority vector), and producing higher-quality images (0.82 vs. 0.73). ITI-GEN frequently generates artifacts, with fewer images containing a person (92.8\% vs. 99\%). For SD3.5, \mitname-\hardprompt\ reduces intersectional biases effectively ($\tau = 0.27$) with 76\% of the steps. Being training-free, it maintains the original model's image quality, unlike ITI-GEN.

\section{Discussion}
\label{sec:discussion}

\noindent\textbf{Role of priority vector.} Incorporating user-defined priorities enables flexible and targeted bias mitigation. For example, in Fig. \ref{fig: applications}(c), the user assigns equal weights to body type, environment, and clothing, prompting the model to mitigate all three biases equally. In contrast, Fig. \ref{fig: applications}(b) illustrates a case where the user prioritizes clothing diversity while assigning lower weight to emotion, focusing the mitigation effort accordingly. This flexibility makes our approach adaptable to a wide range of user goals and fairness requirements.

\noindent\textbf{Accounting for different target distributions.} 
Most fairness methods assume a fixed ideal distribution. In contrast, \modelname\ allows users to define a custom target $D^*$ per bias axis, enabling context-sensitive mitigation. As an experiment, we collect 48 real images of computer programmers, and use this to replace $D^*$ with $D^{real}$ for all biases. Now, re-estimating the \iss, we observe significant differences (Appendix Fig. \ref{fig:realworld}). Notably, the \iss\ for the effect of mitigating clothing on itself flips from +0.88 to -0.79 in Kandinsky, as the ideal distribution of clothing now reflects the skew towards \textit{informal} in the real world, rather than a uniform distribution (see Appendix \ref{sup:real_world}).

\noindent\textbf{Uncovering optimal bias mitigation strategies.} \mitname\ is flexible and supports any set of user-specified bias axes. As shown in Fig. \ref{fig: applications}(a) \& (c), it often achieves effective mitigation in fewer steps than the user-defined threshold. By leveraging inter-axis relations, it identifies optimal strategies. In Fig. \ref{fig: applications}(a), when age and ethnicity are equally prioritized, mitigating ethnicity alone can reduce both due to demographic overlap, and a single intervention meets the threshold. In Fig. \ref{fig: applications}(c) in two mitigation steps, the bias profile progressively aligns with the priority vector (dot product $\tau$ drops: 0.98 → 0.50 → 0.19). Notably, mitigating environment also reduces clothing bias due to strong intersectionality, showing how our method leverages inter-axis relationships for efficient mitigation. Moreover, if \mitname\ fails to reach the desired bias threshold or if mitigating one axis negatively impacts another, it can alert the user to these trade-offs (Fig.~\ref{fig:examples}), enabling informed decision-making.

\noindent\textbf{Extension to Other Approaches.}
We propose a general framework for mitigating intersectional biases in TTI models. As shown in Alg.~\ref{alg:intersectional_mitigation}, our method can be layered on any sequential bias mitigation strategy. At each step, one bias is mitigated, and the intersectionality matrix $\mathbf{S}$ is recomputed, enabling iterative application.

\section{Conclusion}
\label{sec:Conclusion}
We propose \modelname\ to investigate intersectional biases in TTI models. While prior research has explored bias detection and mitigation, to the best of our knowledge, no previous work has focused on understanding how biases influence one another. Unlike \mitname, prior bias mitigation strategies did not account for intersectional impacts. We believe our work enables a more nuanced analysis of bias interactions and supports informed decision-making for AI users and developers, fostering more equitable and transparent AI.



\section{Limitation}
We recognize that biases in generative AI models can cause real-world harm by reinforcing stereotypes and affecting marginalized groups. Our tool aims to help researchers better understand and mitigate these biases—not justify or amplify them. We acknowledge that our analysis depends on dataset and evaluation choices, and encourage future work to build on and refine our approach.



\bibliography{custom}

\appendix

\clearpage
\setcounter{page}{1}
\renewcommand{\thesection}{\Alph{section}}
\setcounter{section}{0}
\appendix
\renewcommand{\thefigure}{A\arabic{figure}}
\setcounter{figure}{0}
\section{Appendix}
\label{sec:sup}

\subsection{Occupation Dataset}
\label{sup:dataset}

Table \ref{tab:fulldataset} shows the occupation prompt templates along each of the bias axes. We consider 26 occupations: \texttt{\small computer programmer, elementary school teacher, librarian, announcer, pharmacist, chef, chemist, police, accountant, architect, lawyer, philosopher, scientist, doctor, nurse, engineer, musician, journalist, athlete, social worker, sales person, politician, farmer, mechanic, firefighter, gardener}.
\begin{table}
  \centering
  \resizebox{\columnwidth}{!}{
  \begin{tabular}{@{}ll@{}}
    \toprule
     \textbf{Axis} & \textbf{Counterfactuals} \\
     \midrule
\multirow{2}{*}{gender bias}  &  A photo of a male \textit{occupation} \\
 &  A photo of a female \textit{occupation} \\
\midrule
\multirow{3}{*}{age bias}  &  A photo of an old \textit{occupation} \\
  &  A photo of a middle-aged \textit{occupation} \\
 &  A photo of a young \textit{occupation} \\
\midrule
\multirow{6}{*}{ethnicity bias} &  A photo of a white \textit{occupation} \\
 &  A photo of a african american \textit{occupation} \\
  &  A photo of an asian \textit{occupation} \\
 &  A photo of a south asian \textit{occupation} \\
  &  A photo of a middle eastern \textit{occupation} \\
  &  A photo of a hispanic \textit{occupation} \\
\midrule
\multirow{3}{*}{bodytype bias} &  A photo of a obese \textit{occupation} \\
  &  A photo of a slim \textit{occupation} \\
  &  A photo of a muscular \textit{occupation} \\
\midrule
\multirow{2}{*}{environment bias}  &  A photo of a \textit{occupation} working indoors \\
 &  A photo of a \textit{occupation} working outdoors \\
\midrule
\multirow{2}{*}{clothing bias}  &  A photo of a \textit{occupation} in formal attire \\
 &  A photo of a \textit{occupation} in informal attire \\
\midrule
\multirow{4}{*}{emotion bias}  &  A photo of a \textit{occupation} who is happy \\
 &  A photo of a sad \textit{occupation} who is sad \\
  &  A photo of a \textit{occupation} who is serious \\
 &  A photo of a \textit{occupation} who is tired \\
\midrule
\multirow{4}{*}{disability bias}  &  A photo of a \textit{occupation} who is fit \\
 &  A photo of a blind \textit{occupation} \\
  &  A photo of a \textit{occupation} with a hearing aid \\
  &  A photo of a \textit{occupation} on a wheelchair \\
    \bottomrule
    \end{tabular}}
    \caption{\textbf{Occupation Dataset.}}
    \label{tab:fulldataset}
\end{table}

\subsection{Models}
\label{sup:models}

We test \modelname\ on several different TTI models. Unless mentioned otherwise, all the settings for image generation are the default settings provided for these models on Huggingface. We use a single Nvidia A40 GPU for all our experiments. We do not measure exact GPU hours as we do not do not train any large scale models, and our usage is dependent on the number of biases we are analyzing or mitigating for each prompt.

For prompts in the Occupation dataset, we use the following models:

\begin{itemize}
    \item Stable Diffusion 1.4: We pick this model because of its compatibility with most bias mitigation tools, like ITI-GEN. We use \textit{CompVis/stable-diffusion-v1-4}.
    \item Stable Diffusion 3.5: This is the latest model in the Stable Diffusion family. We specifically use the \textit{stabilityai/stable-diffusion-3.5-large-turbo} model.
    \item Flux-dev: The Flux series of models is another popular set of open-source TTI models. We use the \textit{black-forest-labs/FLUX.1-dev} variant of this family, and conduct inference with \texttt{guidance scale = 3.5} and \texttt{num inference steps = 30}.
    \item Playground 2.5: This model is trained to produce aesthetically pleasing images. We use \textit{playgroundai/playground-v2.5-1024px-aesthetic}, with \texttt{guidance scale = 3} and \texttt{num inference steps = 50}.
    \item Kandinsky 2.2: We use the \textit{kandinsky-community/kandinsky-2-2-decoder} model, and use the default \texttt{negative prompt = "low quality, bad quality"}.
\end{itemize}

For the TIBET dataset, we use the images already provided in the dataset as is. These images were generated using Stable Diffusion 2.1.

\subsection{VQA}
\label{sup:VQA}

For a given set of images and a set of axes $B$, our goal is to find distributions $D_{B_i}$ for all bias axes $B_i \in B$. In order to find this distribution, we must begin by first identifying the attributes related to $B_i$ in every image of the image set. We use VQA for this process.

For every image in the set, we first start by asking the VQA the question \texttt{\small Is there a person in the image (yes or no)?} for the Occupation prompts dataset. This allows us to filter images where we will be unable to extract bias-related attributes due to low quality generation. For the images that have a person, we have the following set of questions to extract all bias-related attributes:

\begin{itemize}
\item  gender bias:  \texttt{\small What is the gender (male, female) of the person?}
\item  age bias:  \texttt{\small What is the age group (young, middle, old) of the person?}
\item  ethnicity bias:  \texttt{\small What is the ethnicity (white, black, asian, south asian, middle eastern, hispanic) of the person?}
\item  bodytype bias:  \texttt{\small What is the body type (fat, slim, muscular) of the person?}
\item  environment bias:  \texttt{\small What is the environment (indoor, outdoor) of the person?}
\item  clothing bias:  \texttt{\small What is the attire (formal, informal) of the person?}
\item  emotion bias:  \texttt{\small What is the emotion (happy, sad, serious, tired) of the person?}
\item  disability bias:  \texttt{\small Is this person blind (yes or no)?; Is this person wearing a hearing aid (yes or no)?; Is this person on a wheelchair (yes or no)?}
\end{itemize}

Note that all questions are multiple choice. Furthermore, for disability bias, we split the question into three parts, and run each part through the VQA model independently. If none of the parts are answered as `yes', then the person in the image is `fit' and does not have one of those disabilities.

In terms of error rate for robustness, we believe that our MCQ-based VQA approach would yield a lower than 18\% error rate observed in TIBET \cite{chinchure2024tibet}, which uses the same VQA model. Empirically speaking, we observe that our VQA performs near-perfectly on axes such as gender, environment and emotion, but may sometimes return incorrect guesses among other axes in more ambiguous scenarios. As VQA models improve, our method can utilize them in a plug-and-play manner.

\subsection{TIBET Data}
\label{sup:TIBET}

TIBET dataset contains 100 prompts, their biases and relevant counterfactuals, and 48 images for each initial and counterfactual prompt. Because of the dynamic nature of these biases (they vary from prompt to prompt), we use the VQA strategy in the TIBET method instead of our templated questions from above to extract concepts.

\subsection{Bias Mitigation Study}
\label{sup:intro_study}

We conduct a study using ITI-GEN to measure how often a bias mitigation might yield negative effects on other bias axes. We define a negative \isscore\ score ($IS_{xy} < 0$) to suggest that mitigating bias axis $B_x$ reduces the diversity of attributes of axis $B_y$. 

In this study, for all 26 occupations and across all bias axes listed in Table \ref{tab:fulldataset}, we mitigate every bias axis independently. We then compute \iss, where the initial distribution $D_{B_y}^{B_x}$ in equation 3 is replaced by $D_{B_y}^{mit(B_x)}$, which is based on the VQA extracted attributes for bias axis $B_y$ in the newly generated set of images post-mitigation of axis $B_x$ with ITI-GEN. 
This score is defined as:
\begin{equation}
w_{B_y}^{B_x} = W_1(D_{B_y}^{mit(B_x)}, D^*)
\end{equation}
\begin{equation}
IS_{xy}^{mit(x)} = \overline{w}_{B_y}^{\text{init}} - \overline{w}_{B_y}^{B_x} 
\end{equation}

We compute the percentage of $IS_{xy}^{mit(x)}$ for all possible pairs of biases, $B_x$ and $B_y$, where mitigation of $B_x$ led to $IS_{xy}^{mit(x)} < 0$. We find that a substantial number of times, $29.4\%$ of all mitigations, led to a negative effect.

\subsection{Additional prompt-level examples}
\label{sup:TIBET}

We show additional examples of prompt-level intersectional analysis in Fig \ref{fig:moreegs} below. For TIBET, Fig \ref{fig:moreegs}(c) shows how diversifying on an axis like Geography can help diversify the Ethnicity distribution. 

\subsection{Prompt-Modification Based Mitigation (\hardprompt)}
\label{sup:hardprompt}

We use the simplest possible mitigation method: using prompt modification. We choose to use a prompt-modification based method over other methods like ITI-GEN because it gives us the capability to mitigate biases sequentially, store intermediate results, and evaluate the effectiveness of our mitigation algorithm \mitname. Moreover, it is training-free, and can leverage compute optimizations like quantization and Flash Attention to reduce computational costs.

Let us assume we want to mitigate environment bias, and then clothing bias for "nurse". We will assume that our ideal distribution is the uniform distribution across all counterfactuals of each of these axes. The prompt modification process (\hardprompt) works as follows:
\begin{itemize}
    \item Environment bias has two counterfactuals, `\textit{indoor}' and `\textit{outdoor}'. If our total set of images is 48, mitigating it would mean generating 50\% images indoor, and 50\% outdoor. Therefore, during mitigation, our initial prompt \texttt{\small A photo of a nurse} is replaced by a combination of two initial prompts, \texttt{\small A photo of a nurse working \textit{indoors}, A photo of a nurse working \textit{outdoors}}. This is our mitigated model. At this stage, any counterfactual prompt (for \modelname) will also account for this mitigated prompt set. So the gender counterfactuals, at this step, will be  \texttt{\small [A photo of a \textit{male} nurse working indoors, A photo of a \textit{male} nurse working outdoors], [A photo of a \textit{female} nurse working indoors, A photo of a \textit{female} nurse working outdoors]}. 
    \item Next, we want to mitigate clothing bias. Clothing, again, has two counterfactuals: `\textit{formal}' and `\textit{informal}'. Our new initial prompt set will be \texttt{\small A photo of a nurse working indoors dressed \textit{formally}, A photo of a nurse working outdoors dressed \textit{formally}, A photo of a nurse working indoors dressed \textit{informally}, A photo of a nurse working outdoors dressed \textit{informally}}. 
    \item All future mitigation steps will add to these permutations, and an equal number of images are generated for each prompt in the set, such that the total is 48 (or more) images.
\end{itemize}

\subsection{Mitigation Prompts}
\label{sup:mitprompts}

Table \ref{tab:allsd14} contains all the occupation prompts, and their mitigation results, for the SD1.4 model using \mitname-\hardprompt\ and ITI-GEN. Table \ref{tab:allsd35} has all prompts and mitigation results for the SD3.5 experiments. This table also shows cases where we did mitigation based on a priority vector.

\subsection{Validating Mitigation Effect Estimation}
\label{sup:mitigation}

Our approach provides empirical estimates of how a counterfactual-based mitigation strategy may influence an intersectional relationship $B_x \to B_y$ in the form of the \isscore\ score. To validate these estimates, we conduct an experiment where we actually perform mitigation on SD 1.4 using ITI-GEN and SD3.5 using \hardprompt. For all 26 occupations, we consider all intersectional relationships $B_x \to B_y$, and mitigate all $B_x$ independently. To compute the new \isscore\ post mitigation, we replace the initial distribution $D_{B_y}^{B_x}$ in equation 3 with $D_{B_y}^{mit(B_x)}$, which is based on the VQA extracted attributes for bias axis $B_y$ in the newly generated set of images post-mitigation of axis $B_x$ with ITI-GEN. This new score can be defined as:

\begin{equation}
w_{B_y}^{B_x} = W_1(D_{B_y}^{mit(B_x)}, D^*)
\end{equation}
\begin{equation}
IS_{xy}^{mit(x)} = \overline{w}_{B_y}^{\text{init}} - \overline{w}_{B_y}^{B_x} 
\end{equation}

Note that these equations are the same as the ones we used in Section \ref{sup:intro_study}. To quantify the effectiveness of \modelname\, we measure the average correlation between the \isscore\ scores before $IS_{xy}$ and after mitigation $IS_{xy}^{mit(x)}$ across all intersectional relationships $B_x \to B_y$ present for each prompt.

The high correlations (0.65 for ITI-GEN, 0.95 for \hardprompt) suggest that our method effectively estimates the potential impacts of bias interventions without actually doing the mitigation step itself, which can be computationally expensive.

Such empirical guarantees provide users with valuable insights into whether altering bias along a particular dimension will lead to meaningful improvements in fairness across other bias dimensions. By estimating how counterfactual-based interventions influence overall bias scores, our approach helps researchers and practitioners predict the effectiveness of mitigation techniques before full deployment.

\subsection{Using Real World Biases}
\label{sup:real_world}
Other than understanding biases in TTI models
\modelname\ can be used to compare bias dependencies in images generated by Text-to-Image (TTI) models with a reference real-world image distribution. Instead of assuming a uniform distribution as the baseline for bias sensitivity calculations, we consider the empirical distribution of the reference dataset as the initial distribution.

Given a prompt $P$ (e.g., ``A computer programmer''), let $B = [B_1, B_2, ..., B_n]$ represent the set of bias axes (e.g., gender, age, race). For each bias axis $B_y$, we define:
\begin{itemize}
    \item $D_{B_y}^{\text{real}}$: real-world distribution of $B_y$ (from a dataset or observed statistics).
    \item $D_{B_y}^{\text{init}}$: distribution of $B_y$ in TTI-generated images. This is the same as in \modelname.
\end{itemize}

The Wasserstein-1 distance  between real-world and TTI-generated distributions quantifies how far the TTI bias distribution is from real-world data is:
\begin{equation}
    w_{B_y}^{\text{init}} = W_1(D_{B_y}^{\text{init}}, D_{B_y}^{\text{real}})
\end{equation}

To measure the impact of intervening on $B_x$, we compute the post-intervention Wasserstein distance:
\begin{equation}
    w_{B_y}^{B_x} = W_1(D_{B_y}^{B_x}, D_{B_y}^{\text{real}})
\end{equation}
The \isscore\ Score $IS_{xy}$ for the effect of changing $B_x$ on $B_y$ measures the difference between $w_{B_y}^{\text{init}}$ and $ w_{B_y}^{B_x}$ similar to the one calculated in Eq \ref{eqn:metric}.

\begin{figure*}
  \centering
   \includegraphics[width=0.9\linewidth]{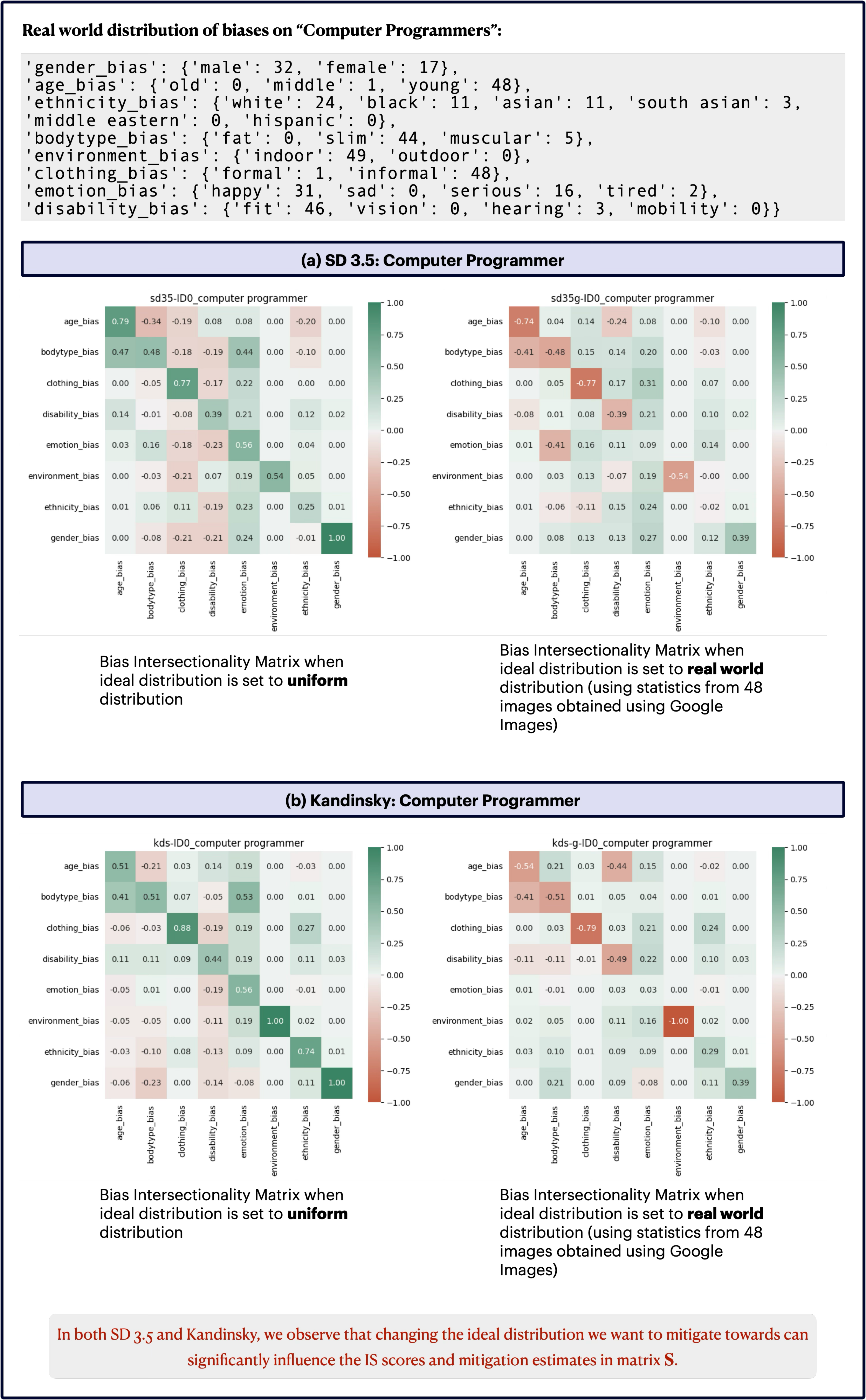}
   \caption{Modifying $D^*$ (ideal distribution) in \modelname\ can have a significant effect on the \isscore\ values.}
   \label{fig:realworld}
\end{figure*}

In our experiment, we obtain a real-world distribution of all biases for a \texttt{\small computer programmer} by sampling 48 images from Google Images. We measure the real world distributions using VQA. Now, by using \modelname\ with the real-world distributions of each bias set to be the ideal distribution $D^*$, we recompute \iss\ as described above. Fig. \ref{fig:realworld} show how significant the effect of this is. Notably, in both SD3.5 and Kandinsky images, we observe that mitigating clothing (to diversify clothing) actually has a negative $IS$ value, as this would move away from our ideal (real-world) distribution of computer programmers mostly wearing informal clothes.

\begin{figure*}
  \centering
   \includegraphics[width=0.9\linewidth]{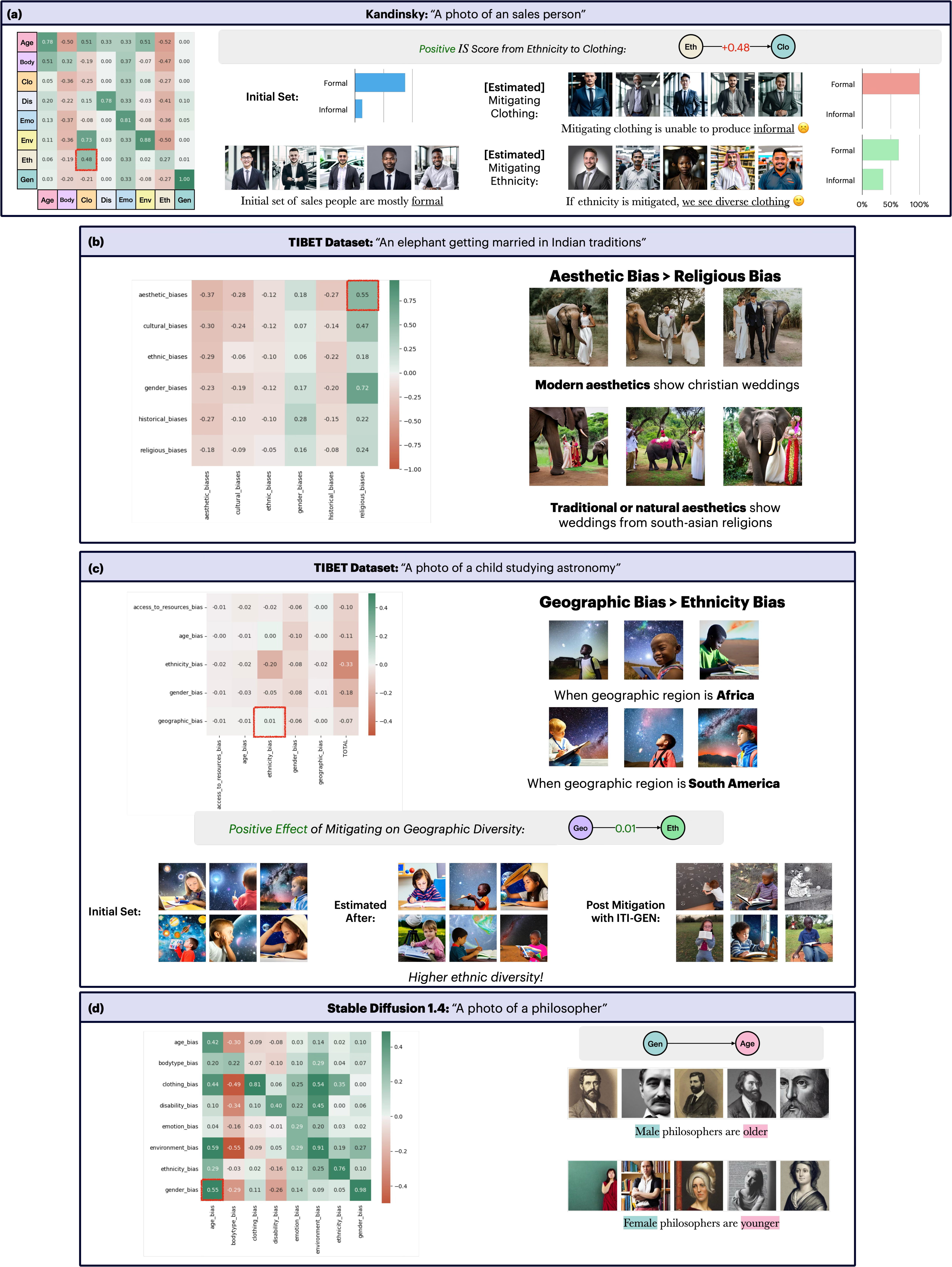}
   \caption{Additional examples on TIBET (b,c) and Occupation prompt (a,d) on prompt-level analysis provided by \modelname.}
   \label{fig:moreegs}
\end{figure*}

\begin{table*}[t]
\small
\rotatebox{90}{\resizebox{1\textwidth}{!}{
\begin{tabular}{lp{0.3\linewidth}|p{0.3\linewidth}rrr|r}
\textbf{Occupation} &
  \textbf{Mitigation Priority Vector} &
  \textbf{Bias Mitigated} &
  \multicolumn{1}{l}{\textbf{\# of Biases}} &
  \multicolumn{1}{l}{\textbf{\# Steps}} &
  \multicolumn{1}{l}{\textbf{MitAmt}} &
  \multicolumn{1}{l}{\textbf{ITI-GEN MitAmt}} \\
  \toprule
computer programmer &
  \{'age\_bias': 0.33, 'emotion\_bias': 0.33, 'ethnicity\_bias': 0.33\} &
  {[}'emotion\_bias', 'ethnicity\_bias', 'age\_bias'{]} &
  3 &
  3 &
  0.46 &
  0.64 \\
elementary school teacher &
  \{'bodytype\_bias': 0.33, 'ethnicity\_bias': 0.33, 'gender\_bias': 0.33\} &
  gender\_bias', 'bodytype\_bias'{]} &
  3 &
  2 &
  0.28 &
  0.51 \\
librarian &
  \{'disability\_bias': 0.25, 'emotion\_bias': 0.25, 'ethnicity\_bias': 0.25, 'gender\_bias': 0.25\} &
  {[}'disability\_bias', 'age\_bias', 'ethnicity\_bias'{]} &
  4 &
  3 &
  0.32 &
  0.60 \\
announcer &
  \{'age\_bias': 0.5, 'environment\_bias': 0.5\} &
  {[}'age\_bias', 'environment\_bias'{]} &
  2 &
  2 &
  0.04 &
  0.18 \\
pharmacist &
  \{'bodytype\_bias': 0.25, 'disability\_bias': 0.25, 'emotion\_bias': 0.25, 'ethnicity\_bias': 0.25\} &
  {[}'age\_bias', 'bodytype\_bias', 'disability\_bias', 'ethnicity\_bias'{]} &
  4 &
  4 &
  0.40 &
  0.77 \\
chef &
  \{'disability\_bias': 0.5, 'emotion\_bias': 0.5\} &
  {[}'emotion\_bias', 'disability\_bias'{]} &
  2 &
  2 &
  0.69 &
  0.82 \\
chemist &
  \{'age\_bias': 0.33, 'clothing\_bias': 0.33, 'emotion\_bias': 0.33\} &
  {[}'emotion\_bias', 'bodytype\_bias'{]} &
  3 &
  2 &
  0.33 &
  0.40 \\
police &
  \{'age\_bias': 0.33, 'emotion\_bias': 0.33, 'ethnicity\_bias': 0.33\} &
  {[}'age\_bias', 'ethnicity\_bias', 'emotion\_bias'{]} &
  3 &
  3 &
  0.39 &
  0.58 \\
accountant &
  \{'age\_bias': 0.25, 'clothing\_bias': 0.25, 'environment\_bias': 0.25, 'ethnicity\_bias': 0.25\} &
  {[}'ethnicity\_bias', 'environment\_bias'{]} &
  4 &
  2 &
  0.28 &
  0.28 \\
architect &
  \{'bodytype\_bias': 0.2, 'emotion\_bias': 0.2, 'environment\_bias': 0.2, 'ethnicity\_bias': 0.2, 'gender\_bias': 0.2\} &
  {[}'emotion\_bias', 'gender\_bias', 'bodytype\_bias'{]} &
  5 &
  3 &
  0.34 &
  0.51 \\
lawyer &
  \{'emotion\_bias': 0.5, 'gender\_bias': 0.5\} &
  {[}'emotion\_bias'{]} &
  2 &
  1 &
  0.33 &
  0.33 \\
philosopher &
  \{'clothing\_bias': 0.5, 'disability\_bias': 0.5\} &
  {[}'clothing\_bias'{]} &
  2 &
  1 &
  0.33 &
  0.56 \\
scientist &
  \{'age\_bias': 0.5, 'emotion\_bias': 0.5\} &
  {[}'bodytype\_bias', 'disability\_bias'{]} &
  2 &
  2 &
  0.38 &
  0.49 \\
doctor &
  \{'age\_bias': 0.2, 'bodytype\_bias': 0.2, 'clothing\_bias': 0.2, 'ethnicity\_bias': 0.2, 'gender\_bias': 0.2\} &
  {[}'gender\_bias', 'bodytype\_bias'{]} &
  5 &
  2 &
  0.28 &
  0.45 \\
nurse &
  \{'bodytype\_bias': 0.33, 'clothing\_bias': 0.33, 'environment\_bias': 0.33\} &
  {[}'gender\_bias', 'environment\_bias'{]} &
  3 &
  2 &
  0.31 &
  0.42 \\
engineer &
  \{'age\_bias': 0.5, 'disability\_bias': 0.5\} &
  {[}'bodytype\_bias', 'disability\_bias'{]} &
  2 &
  2 &
  0.53 &
  0.61 \\
musician &
  \{'age\_bias': 0.2, 'disability\_bias': 0.2, 'emotion\_bias': 0.2, 'ethnicity\_bias': 0.2, 'gender\_bias': 0.2\} &
  {[}'age\_bias', 'gender\_bias', 'ethnicity\_bias', 'disability\_bias'{]} &
  5 &
  4 &
  0.31 &
  0.67 \\
journalist &
  \{'emotion\_bias': 0.5, 'ethnicity\_bias': 0.5\} &
  {[}'emotion\_bias', 'ethnicity\_bias'{]} &
  2 &
  2 &
  0.33 &
  0.39 \\
athlete &
  \{'age\_bias': 0.25, 'bodytype\_bias': 0.25, 'disability\_bias': 0.25, 'gender\_bias': 0.25\} &
  {[}'gender\_bias', 'age\_bias', 'disability\_bias'{]} &
  4 &
  3 &
  0.30 &
  0.56 \\
social worker &
  \{'bodytype\_bias': 0.33, 'ethnicity\_bias': 0.33, 'gender\_bias': 0.33\} &
  {[}'emotion\_bias', 'bodytype\_bias'{]} &
  3 &
  2 &
  0.24 &
  0.18 \\
sales person &
  \{'age\_bias': 0.5, 'ethnicity\_bias': 0.5\} &
  {[}'age\_bias', 'ethnicity\_bias'{]} &
  2 &
  2 &
  0.22 &
  0.48 \\
politician &
  \{'age\_bias': 0.33, 'bodytype\_bias': 0.33, 'ethnicity\_bias': 0.33\} &
  {[}'ethnicity\_bias'{]} &
  3 &
  1 &
  0.34 &
  0.47 \\
farmer &
  \{'age\_bias': 0.5, 'ethnicity\_bias': 0.5\} &
  {[}'gender\_bias', 'ethnicity\_bias'{]} &
  2 &
  2 &
  0.27 &
  0.46 \\
mechanic &
  \{'bodytype\_bias': 0.2, 'clothing\_bias': 0.2, 'emotion\_bias': 0.2, 'environment\_bias': 0.2, 'ethnicity\_bias': 0.2\} &
  {[}'bodytype\_bias', 'emotion\_bias', 'clothing\_bias', 'age\_bias', 'ethnicity\_bias'{]} &
  5 &
  5 &
  0.37 &
  0.74 \\
firefighter &
  \{'age\_bias': 0.25, 'clothing\_bias': 0.25, 'environment\_bias': 0.25, 'gender\_bias': 0.25\} &
  {[}'age\_bias', 'clothing\_bias', 'gender\_bias'{]} &
  4 &
  3 &
  0.30 &
  0.67 \\
gardener &
  \{'age\_bias': 0.33, 'environment\_bias': 0.33, 'ethnicity\_bias': 0.33\} &
  {[}'environment\_bias', 'ethnicity\_bias'{]} &
  3 &
  2 &
  0.33 &
  0.63 \\
  \bottomrule
\end{tabular}}}
\caption{All 26 occupational prompts and their associated biases that were mitigated (randomly selected), on Stable Diffusion 1.4. We show the priority vector, along with the actual list of biases that were mitigated, and the corresponding number of steps. Finally, we also mention MitAmt using our approach and ITI-GEN. The aggregate results of this table are in the main paper, Table \ref{tab:mitigationres}.}
  \label{tab:allsd14}
\end{table*}
\begin{table*}[t]
\small
\rotatebox{90}{\resizebox{1.4\textwidth}{!}{
\begin{tabular}{lllccc}
\textbf{Occupation} &
  \textbf{Mitigation Priority Vector} &
  \textbf{Bias Mitigated} &
  \textbf{\# of Biases} &
  \textbf{\# Steps} &
  \textbf{MitAmt} \\
  \toprule
announcer &
  clothing\_bias,0.5;gender\_bias,0.5 &
  {[}'gender\_bias'{]} &
  2 &
  1 &
  0.13 \\
politician &
  clothing\_bias,0.5;environment\_bias,0.5 &
  {[}'ethnicity\_bias', 'environment\_bias'{]} &
  2 &
  2 &
  0.40 \\
musician &
  age\_bias,0.33;ethnicity\_bias,0.33 &
  {[}'ethnicity\_bias'{]} &
  2 &
  1 &
  0.20 \\
mechanic &
  age\_bias,0.33;bodytype\_bias,0.33;clothing\_bias,0.33 &
  {[}'age\_bias', 'bodytype\_bias', 'clothing\_bias', 'disability\_bias'{]} &
  3 &
  4 &
  0.28 \\
nurse &
  clothing\_bias,0.33;environment\_bias,0.33;bodytype\_bias,0.33 &
  {[}'bodytype\_bias', 'environment\_bias'{]} &
  3 &
  2 &
  0.19 \\
gardener &
  age\_bias,0.33;bodytype\_bias,0.33;gender\_bias,0.33 &
  {[}'gender\_bias'{]} &
  3 &
  1 &
  0.28 \\
sales &
  clothing\_bias,0.33;environment\_bias,0.33;gender\_bias,0.33 &
  {[}'environment\_bias', 'gender\_bias'{]} &
  3 &
  2 &
  0.22 \\
journalist &
  gender\_bias,0.25;clothing\_bias,0.25;emotion\_bias,0.25;ethnicity\_bias,0.25 &
  {[}'ethnicity\_bias'{]} &
  4 &
  1 &
  0.35 \\
engineer &
  bodytype\_bias,0.33;gender\_bias,0.33;age\_bias,0.33;clothing\_bias,0.33 &
  {[}'age\_bias', 'clothing\_bias', 'gender\_bias'{]} &
  4 &
  3 &
  0.28 \\
computer &
  age\_bias,0.33;bodytype\_bias,0.33;clothing\_bias,0.33;disability\_bias,0.33 &
  {[}'disability\_bias', 'bodytype\_bias', 'clothing\_bias'{]} &
  4 &
  3 &
  0.34 \\
athlete &
  age\_bias,0.33;bodytype\_bias,0.33;clothing\_bias,0.33;disability\_bias,0.33;gender\_bias,0.33 &
  {[}'disability\_bias', 'bodytype\_bias', 'age\_bias', 'gender\_bias', 'clothing\_bias'{]} &
  5 &
  5 &
  0.32 \\
doctor &
  age\_bias,0.33;ethnicity\_bias,0.33;clothing\_bias,0.33;emotion\_bias,0.33;gender\_bias,0.33 &
  {[}'ethnicity\_bias', 'age\_bias', 'disability\_bias', 'gender\_bias', 'bodytype\_bias'{]} &
  5 &
  5 &
  0.29 \\
teacher &
  bodytype\_bias,0.33;emotion\_bias,0.33;gender\_bias,0.33;age\_bias,0.33;clothing\_bias,0.33 &
  {[}'environment\_bias', 'clothing\_bias', 'bodytype\_bias'{]} &
  5 &
  3 &
  0.28 \\
chef &
  bodytype\_bias,0.33;emotion\_bias,0.33;gender\_bias,0.33;age\_bias,0.33;clothing\_bias,0.33 &
  {[}'environment\_bias', 'age\_bias', 'bodytype\_bias', 'gender\_bias', 'emotion\_bias'{]} &
  5 &
  5 &
  0.28 \\
  \midrule
librarian &
  environment\_bias,0.3;clothing\_bias,0.7 &
  {[}'environment\_bias', 'clothing\_bias'{]} &
  2 &
  2 &
  0.15 \\
announcer &
  emotion\_bias,0.6;ethnicity\_bias,0.4 &
  {[}'ethnicity\_bias', 'emotion\_bias'{]} &
  2 &
  2 &
  0.50 \\
journalist &
  clothing\_bias,0.8;emotion\_bias,0.2 &
  {[}'ethnicity\_bias'{]} &
  2 &
  1 &
  0.26 \\
accountant &
  clothing\_bias,0.20;environment\_bias,0.15;gender\_bias: 0.65 &
  {[}'clothing\_bias', 'ethnicity\_bias', 'emotion\_bias'{]} &
  3 &
  3 &
  0.25 \\
sales person &
  age\_bias,0.5;gender\_bias,0.3;bodytype\_bias,0.2 &
  {[}'gender\_bias', 'age\_bias'{]} &
  3 &
  2 &
  0.29 \\
  \bottomrule
\end{tabular}}}
\caption{Selected occupational prompts and their associated biases that were mitigated (manually selected), on Stable Diffusion 3.5. We show the priority vector, along with the actual list of biases that were mitigated, and the corresponding number of steps. Finally, we also mention MitAmt using our algorithm. The aggregate results of this table are in the main paper, Table \ref{tab:mitigationres}. The bottom section of the table has examples where the priority vector was weighted.}
  \label{tab:allsd35}
\end{table*}


\end{document}